# ArchiLense: A Framework for Quantitative Analysis of Architectural Styles Based on Vision Large Language Models

**Jing Zhong[1,*], Jun Yin[1](🖂), Peilin Li[2], Pengyu Zeng[3], Miao Zang[4], Ran Luo[5], Shuai Lu[6]**

**Abstract.** Architectural cultures across regions are characterized by stylistic diversity, shaped by historical, social, and technological contexts in addition to geographical conditions. Understanding architectural styles requires the ability to describe and analyze the stylistic features of different architects from various regions through visual observations of architectural imagery. However, traditional studies of architectural culture have largely relied on subjective expert interpretations and historical literature reviews, often suffering from regional biases and limited explanatory scope.

To address these challenges, this study proposes three core contributions: (1) We construct a professional architectural style dataset named ArchDiffBench, which comprises 1,765 high-quality architectural images and their corresponding style annotations, collected from different regions and historical periods. (2) We propose ArchiLense, an analytical framework grounded in Vision-Language Models and constructed using the ArchDiffBench dataset. By integrating advanced computer vision techniques, deep learning, and machine learning algorithms, ArchiLense enables automatic recognition, comparison, and precise classification of architectural imagery, producing descriptive language outputs that articulate stylistic differences. (3) Extensive evaluations show that ArchiLense achieves strong performance in architectural style recognition, with a 92.4% consistency rate with expert annotations and 84.5% classification accuracy, effectively capturing stylistic distinctions across images.

The proposed approach transcends the subjectivity inherent in traditional analyses and offers a more objective and accurate perspective for comparative studies of architectural culture.

---

[1]Jun Yin (🖂)
Shenzhen International Graduate School, Tsinghua University, Shenzhen, China
e-mail: shuai.lu@sz.tsinghua.edu.cn

[2]National University of Singapore

[*]Equal contribution.

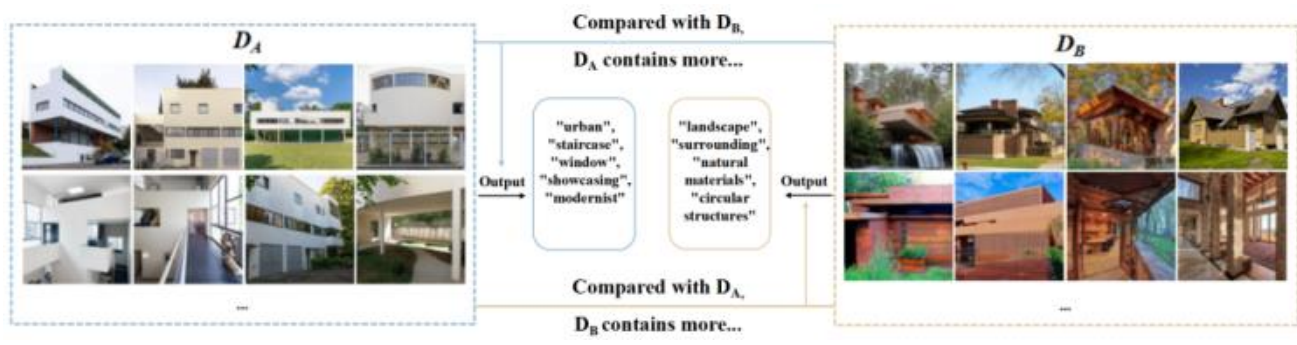

**Fig. 1** Graphic Abstract

# 1 Introduction

Recent advances in machine learning have led to a wide range of studies in sustainable architecture [11,13-5, 37, 46], energy-aware design [5, 8, 17-19], multimodal generation [23-25, 43], visual enhancement [30, 34, 44], architectural performance optimization [15-18, 31] and human-AI interaction [4, 11, 28, 33, 36, 39].Among these, architectural style analysis remains a critical yet relatively underexplored domain.

Systematic analysis and quantitative comparison of architectural styles across regions and historical periods are of critical academic and practical importance to architectural design and cultural studies [27]. Architectural styles reflect not only regional culture and social transformation but also spatial logic and aesthetic ideologies shaped by historical contexts [1]. Therefore, such investigations help reveal underlying evolutionary mechanisms and offer new theoretical foundations for architectural discourse and innovation in contemporary design approaches.

Traditional studies of architectural styles have primarily relied on expert interpretation and historical-visual analysis [20]. However, these approaches are inherently constrained by subjectivity and cognitive limitations, often leading to inconsistencies in classification. Moreover, manual analytical methods lack scalability when dealing with large-scale and heterogeneous datasets. In recent years, with the increasing integration of architectural cultures across regions, traditional methods have struggled to provide reproducible and quantitative comparative frameworks, posing significant challenges to the rigor of architectural style research. Advances in computer vision, deep learning, and natural language processing have shifted architectural style research toward automated and data-driven paradigms.Machine learning enables efficient extraction of stylistic features from large-scale datasets, outperforming traditional expert-based methods. Meanwhile, the rise of Vision Large Language Models (VLLMs) further enhances multimodal analysis, supporting automatic classification and descriptive articulation of architectural styles—laying the groundwork for a systematic and integrated analytical framework.

Accordingly, we proposes ArchiLense, a novel framework based on VLLMs to quantitatively analyze architectural style variations. By integrating deep learning and natural language processing, ArchiLense enables automatic recognition and linguistic articulation of stylistic differences in architectural imagers : (1) VLLM extracts visual features and generates candidate style descriptions; (2) A ranking module selects the most distinctive and expressive descriptions; (3) Selected descriptions guide a text-to-image generation model to synthesize architectural images, which are validated via expert evaluation. Through ArchiLense, we are able to enrich academic inquiry and applications in architectural style studies.

## 2 Related works

### *2.1 Research on Architectural Style*

Architectural style research is a core area in architecture, art history, and heritage studies, centers on identifying design features across historical periods, regions, and movements [9].

Traditional architectural style studies often rely on literature review and expert visual assessment, which, though insightful, are subjective and inefficient at scale. Recent advancements in computer vision have enabled data-driven methods, such as CNN-based models that extract stylistic features and classify large-scale architectural images [27]. However, these approaches are typically confined to classification tasks and lack mechanisms to analyze stylistic evolution or perform cross-style transfer. Additionally, conventional models emphasize low-level features, limiting their ability to capture semantic nuances in architectural styles.

### *2.2 Application of Visual Large Language Model (VLLM)*

In recent years,Vision Large Language Models (VLLMs) represent a major advancement in multimodal AI, combining computer vision and natural language processing for enhanced semantic reasoning.

While VLLMs like CLIP and ViT-GPT perform well in cross-modal tasks, their use in architecture remains limited. Recent work has leveraged CLIP for architectural image-text alignment, such as improved Chinese image labeling via CLIP-enhanced DreamBooth [2], and diffusion models for style transfer and prompt-based generation [10]. However, most studies focus on image synthesis, with limited attention to style analysis and quantitative comparison.

## 3 Dataset

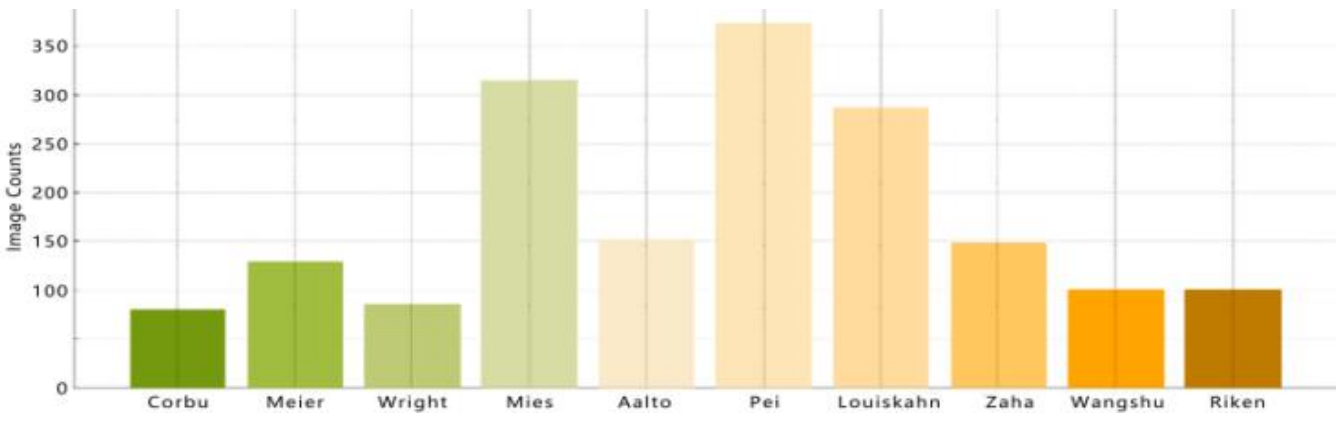

**Fig. 2** The Distribution of Each Category of Images In Dataset

To systematically analyze architectural styles, we developed ArchDiffBench, a dataset integrating expert evaluations and textual analysis. The images were sourced from ArchDaily® and gooood®, two major global repositories of architectural design. ArchDiffBench comprises 1,765 images across various periods and regions, encompassing facades, interiors, and exteriors to ensure visual richness. Images are categorized into 10 groups by architect identity. To analyze stylistic relationships, 81 paired subsets were constructed for inter-group comparison. Figure 2 illustrates the category-wise image distribution.

# 4 Methodology

Inspired by Visdiff [22], we developed the ArchiLense framework, a two-stage analysis system designed for style feature extraction and quantification. As shown in Figure 3, the framework consists of a Style Extractor, which generates discriminative textual descriptions from randomly sampled images, and a Style Evaluator, which validates and quantitatively assesses these descriptions in style differentiation tasks. This approach addresses the limitations of directly training neural networks on complex architectural imagery, where general-purpose models often struggle to balance encoding efficiency and predictive accuracy.

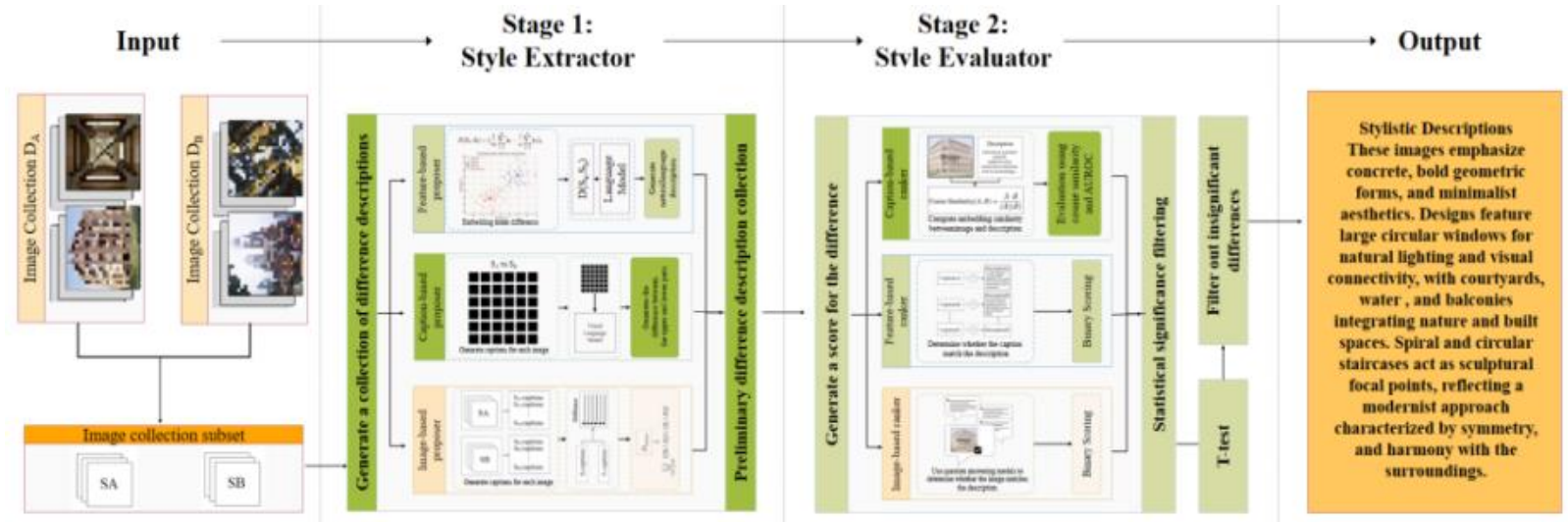

**Fig. 3** ArchiLense Framework

## *4.1 Style Extractor*

The Style Extractor aims to identify distinct stylistic features between two architectural groups $D_A$ and $D_B$ while ensuring the extracted traits remain group-specific and not easily generalized. To explore feasible implementations, we tested three VLM-based approaches: (1) Image-based Analysis: Gridded images from both groups are input to VLMs (e.g., GPT-4V) to identify differences in façade design, geometry, materials, and spatial configuration. (2) Embedding Difference Computation: Style embeddings from each group are averaged and subtracted. The resulting vector is input into a language model (e.g., BLIP-2) for textual style descriptions. (3) Text-based Description Comparison: VLMs generate detailed text for each image, while GPT-4 compares texts between groups to produce stylistic difference descriptions.

Experimental results show that Text-based descriptions more effectively capture architect-specific styles, surpassing visual-only methods. Hence, it is chosen as the primary extraction strategy, with others as baselines..

### *4.2 Style Evaluator*

The Style Extractor's output can be invalid or redundant. Therefore, A style discriminator is used to assess the description's validity and quantify its ability to distinguish architectural styles. It calculates the differentiation fraction $S_y$ of the style description y and sorts it as follows:

$$S_y = \sum_{x \in D_A} \theta(x, y) - \sum_{x \in D_B} \theta(x, y)$$

θ(x,y)：The degree of match between image x and style description y.
We explored three computational methods to evaluate $\theta(x,y)$：(1) VQA-based Matching: LLaVA-1.5's strong multimodal reasoning enables us to classify whether an image x aligns with a style description y in a binary manner: “Does this image match the description?” (Yes/No). (2) Caption-to-Description Matching: Each image is captioned by BLIP-2 to obtain *c*, which is evaluated by Vicuna-1.5 through QA($c$,*y*) for alignment with the style description. The task adopts a (Yes/No) classification format. (3) Embedding Similarity Computation: CLIP ViT-G/14 generates vector representations for both the image *x* and the textual description *y*. The semantic alignment between the two is then quantified using cosine similarity, computed as follows:

$$\theta(x, y) = \frac{e_x \cdot e_y}{\|e_x\| \|e_y\|}$$

As the value is continuous, AUROC is used to assess the discriminative power of style descriptions between $D_A$ and $D_B$.

The embedding-based evaluator shows superior accuracy and efficiency, and is thus used as the main evaluation method, with others as baselines. Additionally, a t-test (significance level 0.05) is conducted on $\theta(x,y)$ distributions to exclude statistically insignificant descriptions.

## 5 Experiment Result

This section presents the feature analysis results of ArchiLense, trained on ArchDiffBench, focusing on: (1) generation of architect-specific stylistic descriptions; (2) evaluation of their effectiveness in capturing architectural style differences; (3) application of these texts in Text-to-Image Generation to synthesize representative architectural images, validated via expert review.

## *5.1 Discriminative Effectiveness and Cognitive Consistency of Style Descriptions*

To evaluate the effectiveness of ArchiLense-generated style descriptions, we assessed both their discriminative power and alignment with architectural design cognition. Style descriptions generated from image pairs reflect differences in design approach, form, materials, and typology. Evaluated using AUROC, these descriptions are ranked by their effectiveness in distinguishing architectural styles. Despite the absence of architect identities, the outputs align well with established design knowledge. Word clouds of key terms further confirm that ArchiLense results are both discriminative and cognitively consistent with expert understanding.

| Select two options | | | | |
|---|---|---|---|---|
| | 1 | corbu | 6 | I_M_Pei |
| | 2 | meier | 7 | Louiskahn |
| | 3 | wright | 8 | zaha_hadid |
| | 4 | mies | 9 | wangshu |
| | 5 | Alvar_Aalto | 10 | riken_yamamoto |

zaha_hadid

| DA vs DB | hypothesis | score1 | score2 | diff | t_stat | p_value | auroc | correct_delta |
|---|---|---|---|---|---|---|---|---|
| zaha_hadid vs riken_yamamoto | 1."The depiction of a building with an emphasis on curvilinear forms and futuristic aesthetics." (0.959) | 0.382 | 0.258 | 0.124 | TRUE | 2.890 | 0.959 | 0.817 |
| | 2."A strikingly modern building with sleek curved facade, providing a sense of movement and dynamism to the static structure." (0.929) | 0.361 | 0.267 | 0.094 | TRUE | 5.637 | 0.929 | 0.654 |
| | 3."A futuristic building featuring an aerodynamic design, presented via an artist's rendering, located in an urban setting." (0.925) | 0.332 | 0.237 | 0.094 | TRUE | 1.704 | 0.925 | 0.666 |
| | 4."An exterior image of a building demonstrating cutting-edge architecture with a sweeping curved design, set amidst a bustling cityscape." (0.922) | 0.337 | 0.227 | 0.110 | TRUE | 1.089 | 0.921 | 0.660 |

**Fig. 4** Using Riken and Zaha Hadid as representative cases, we conducted 81 pairwise comparisons between their respective subgroups in the dataset. Each run involved random sampling of 20 images per group.

**Functional:** Keyword extraction effectively captures stylistic features, as shown in Figure 5 word clouds. CLIP and GPT-4 distinguish between Meier's minimalist ("white", "window") and Yamamoto's urban-functional style ("cityscape", "marble").

**Accuracy:** Model outputs align with architectural cognition, proving its reliability and value. For example, Le Corbusier ("staircase") and Mies van der Rohe ("glass") reflect their design philosophies. Wang Shu ("wooden") captures traditional-modern synthesis, while Aalto ("forest") express organic and monumental themes respectively.

**Generalization:** The model robustly identifies diverse architectural styles—across regions and periods—without predefined labels, enabling objective, bias-free classification.

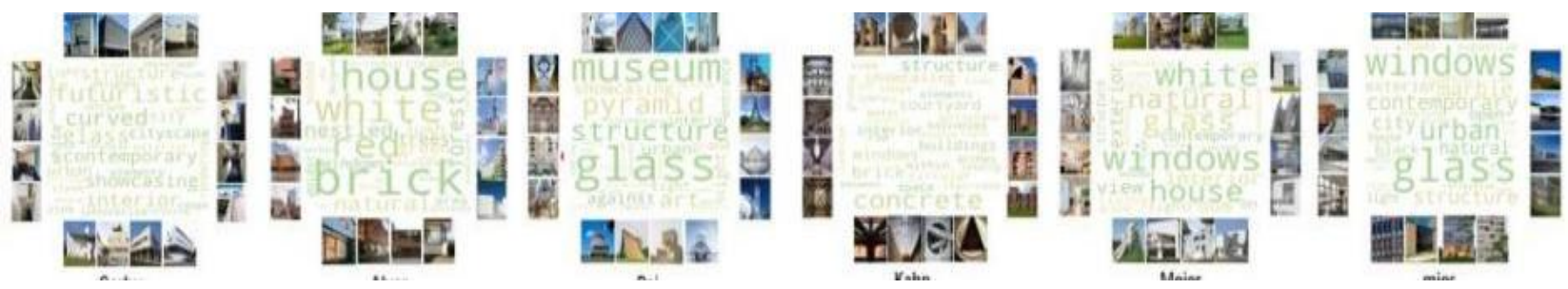

**Fig. 5** Architectural Projects And Corresponding Word Cloud Map

## *5.2 Quantitative Validation of Stylistic Difference Descriptions*

### 5.2.1 T-test and Significance Analysis

An experiment using 270 style descriptions from 9 architect groups evaluated stylistic distinctiveness via Score 1 and Score 2 comparisons, followed by t-tests. As shown in Figure 6, over 80% of the descriptions achieved statistical significance ($p < 0.05$), confirming their effectiveness in differentiating architectural styles.

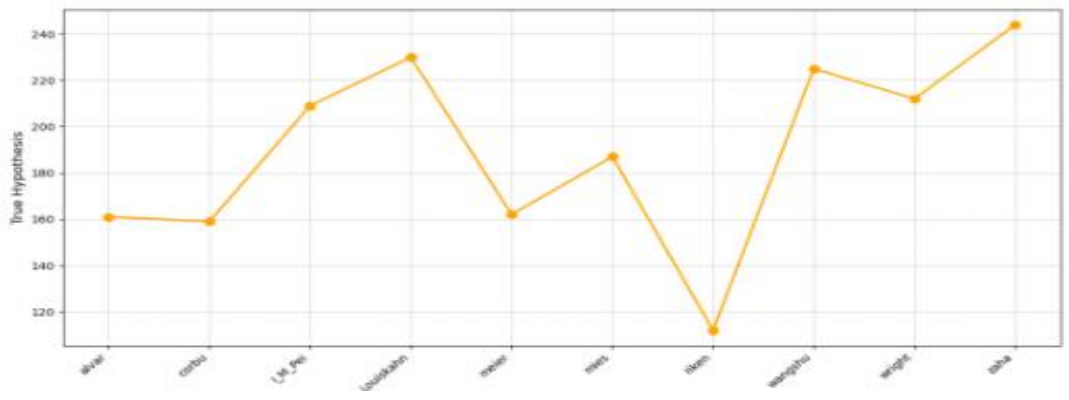

**Fig. 6** The number of descriptions T-tested as "true" can reflect the degree to which an architect's style is unique.

### 5.2.2 Data Validation Visualization

To validate the feature extraction and separation capability of ArchiLense, pairwise comparisons were conducted, visualized through Histogram, KDE, and Boxplot. As shown in Figure 7, the model's performance is visualized across multiple evaluation metrics. The left histogram compares cosine similarities within Groups A and B: Group A's concentrated distribution indicates robust feature consistency, while Group B's dispersion suggests sensitivity to stylistic diversity. Kernel density curves further highlight ArchiLense's balance between consistency and differentiation. The right-side boxplots show that Group A yields higher medians and narrower IQRs, reflecting stable representations, whereas Group B exhibits greater variability and outliers, demonstrating responsiveness to complex stylistic variations.

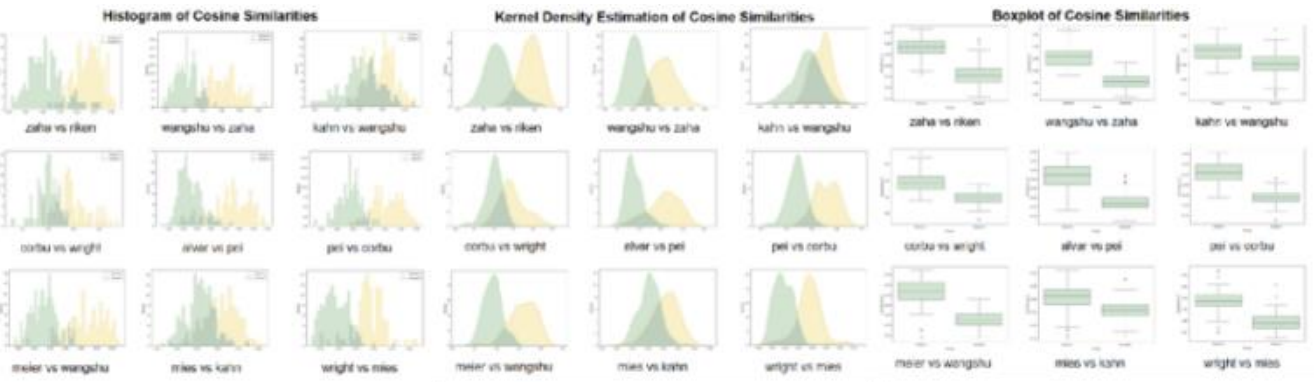

**Fig. 7** Cosine Similarity Distribution and Feature Representation Analysis

### 5.2.3 Architectural Stylistic Difference Attributes

Figure 8 shows textual similarities among architects, with orange indicating high similarity and blue indicating low. Meier, Mies, and Louis Kahn exhibit strong stylistic alignment (0.995), suggesting shared modernist tendencies. Aalto and Wang Shu (0.95) also show notable similarity, possibly indicating stylistic continuity. In contrast, architects with more blue regions tend to express individualistic styles, while orange-dominant patterns may reflect shared academic or traditional influences.

| | alvar | corbu | I_M_Pei | Louiskahn | meier | mies | riken | wangshu |
|---|---|---|---|---|---|---|---|---|
| alvar | 1 | 0.98 | 0.97 | 0.985 | 0.99 | 0.98 | 0.975 | 0.99 |
| corbu | 0.98 | 1 | 0.945 | 0.96 | 0.975 | 0.97 | 0.96 | 0.965 |
| I_M_Pei | 0.97 | 0.945 | 1 | 0.97 | 0.965 | 0.975 | 0.97 | 0.975 |
| Louiskahn | 0.985 | 0.96 | 0.97 | 1 | 0.985 | 0.98 | 0.97 | 0.98 |
| meier | 0.99 | 0.975 | 0.965 | 0.985 | 1 | 0.995 | 0.985 | 0.99 |
| mies | 0.98 | 0.97 | 0.975 | 0.98 | 0.995 | 1 | 0.98 | 0.975 |
| riken | 0.975 | 0.96 | 0.97 | 0.97 | 0.985 | 0.98 | 1 | 0.985 |
| wangshu | 0.99 | 0.965 | 0.975 | 0.98 | 0.99 | 0.975 | 0.985 | 1 |
| wright | 0.97 | 0.95 | 0.965 | 0.975 | 0.985 | 0.97 | 0.975 | 0.96 |
| zaha | 0.965 | 0.94 | 0.97 | 0.98 | 0.98 | 0.975 | 0.97 | 0.95 |

**Fig. 8** Architectural Style Similarity and Clustering Visualization

## *5.3 Feature Style Images Verified by Expert Evaluation*

To assess model performance, 20 architecture PhD students and 5 senior architects rated the consistency between generated images and description texts for 10 architects on a 1 – 5 scale (1 = inconsistent, 5 = fully consistent). Participants matched text-generated images to one of 10 architects. ArchiLense achieved a mean score of 4.62 and an 84.5% matching accuracy, demonstrating strong stylistic reliability and interpretability.

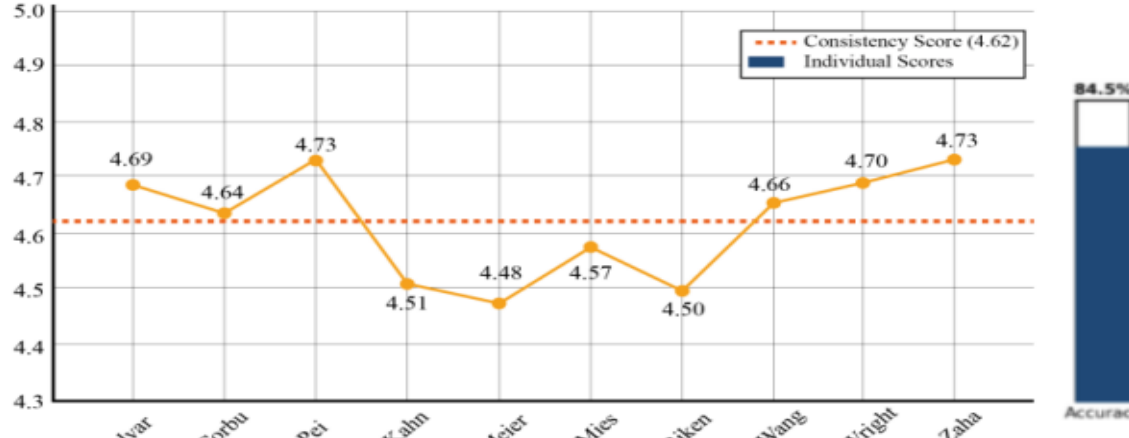

**Fig. 9** Architectural Style Cosine Similarity Matrix

## 6 Discussion

ArchiLense demonstrates the potential of vision-language models for architectural style analysis. However, current limitations include a dataset biased toward modern Western styles and limited analysis of spatial authenticity. Future work will broaden the dataset's diversity and incorporate spatially aware evaluation methods, alongside deeper theoretical engagement, to improve generalizability and interpretability across cultural contexts.

## 7 Conclusion

Although still preliminary, ArchiLense suggests that vision-language models may help improve the efficiency and objectivity of architectural style analysis. By translating visual features into structured language, it provides a potential basis for quantitative and cross-cultural comparisons. Current limitations in dataset diversity and representativeness point to the need for broader coverage and deeper theoretical engagement. With further refinement and expansion, ArchiLense may serve as a useful tool for supporting future research in architectural historiography, style interpretation, and cross-cultural understanding in heritage studies.